\documentclass[conference]{IEEEtran}
\IEEEoverridecommandlockouts
\usepackage{cite}
\usepackage{amsmath,amssymb,amsfonts}
\usepackage{algorithmic}
\usepackage{graphicx}
\usepackage{textcomp}
\usepackage{xcolor}
\usepackage{subcaption}
\usepackage{draftwatermark}

\usepackage{color}
\usepackage[markup=bfit, deletedmarkup=sout, authormarkup=brackets]{changes}
\definechangesauthor[name={Filipe}, color=teal]{FG}
\definechangesauthor[name={Matej}, color=orange]{MH}

\def\BibTeX{{\rm B\kern-.05em{\sc i\kern-.025em b}\kern-.08em
    T\kern-.1667em\lower.7ex\hbox{E}\kern-.125emX}}

\begin{document}

\SetWatermarkAngle{0}
\SetWatermarkColor{black}
\SetWatermarkLightness{0.5}
\SetWatermarkFontSize{10pt}
\SetWatermarkVerCenter{20pt}
\SetWatermarkText{\parbox{30cm}{%
\centering This is the final version of the manuscript that appeared in \\
\centering Proceedings of the 2019 Joint IEEE 9th International Conference on Development and Learning and Epigenetic Robotics (ICDL-EpiRob), \\
\centering Engineer’s House Conference Centre, Oslo, Norway, August 19-22, 2019, pp. 113-114.  (C) IEEE.}}

\title{The homunculus for proprioception: Toward learning the representation of a humanoid robot's joint space using self-organizing maps\\
{\footnotesize \textsuperscript{*}}
\thanks{This work was supported by the Czech Science Foundation under Project GA17-15697Y. F.G. was also partly supported by the Czech Technical University in Prague, grant No. SGS18/138/OHK3/2T/13}
}

\author{\IEEEauthorblockN{Filipe Gama}
\IEEEauthorblockA{\textit{Dpt. of Cybernetics, Faculty of Electrical Engineering} \\
\textit{Czech Technical University in Prague}\\
Prague, Czech Republic \\
filipe.gama@fel.cvut.cz}
\and
\IEEEauthorblockN{Matej Hoffmann}
\IEEEauthorblockA{\textit{Dpt. of Cybernetics, Faculty of Electrical Engineering} \\
\textit{Czech Technical University in Prague}\\
Prague, Czech Republic \\
matej.hoffmann@fel.cvut.cz}
}

\maketitle

\begin{abstract}
In primate brains, tactile and proprioceptive inputs are relayed to the somatosensory cortex which is known for somatotopic representations, or, ``homunculi''. Our research centers on understanding the mechanisms of the formation of these and more higher-level body representations (body schema) by using humanoid robots and neural networks to construct models. We specifically focus on how spatial representation of the body may be learned from somatosensory information in self-touch configurations. In this work, we target the representation of proprioceptive inputs, which we take to be joint angles in the robot. The inputs collected in different body postures serve as inputs to a Self-Organizing Map (SOM) with a  2D lattice on the output. With unrestricted, all-to-all connections, the map is not capable of representing the input space while preserving the topological relationships, because the intrinsic dimensionality of the body posture space is too large. Hence, we use a method we developed previously for tactile inputs (Hoffmann, Straka et al. 2018) called MRF-SOM, where the Maximum Receptive Field of output neurons is restricted so they only learn to represent specific parts of the input space. This is in line with the receptive fields of neurons in somatosensory areas representing proprioception that often respond to combination of few joints (e.g. wrist and elbow). 

\end{abstract}


\section{Introduction}

Somatosensory inputs are constituted by tactile and proprioceptive ones and are first processed by the primary somatosensory cortex (SI) in the primate brain, in particular by Brodmann areas 3b and 3a respectively. The representations are somatotopic, resembling the structure of the human body (hence the term homunculus or ``little man''), yet distorted. These map-like representations are a result of genetic predispositions and experience both before and after birth. For the learning of such maps, self-organizing (Kohonen) maps (SOMs) seem a good candidate because of their topology preservation property. In \cite{HoffmannStraka2018} we used tactile stimulation on the body of a humanoid robot and introduced a modification of SOM that we called MRF-SOM where the Maximum Receptive Field of output neurons is restricted so they only learn to represent specific parts of the input space. The MRF settings were chosen to mimic the genetic predispositions and ensuring the gross layout of the output layer similar to that of area 3b. Pugach et al.~\cite{Pugach2015} have also employed SOM to learn the representation of a tactile surface. 

In \cite{HoffmannBednarova2016} we focused on the representation of proprioceptive inputs, with the additional simplification that only joint angles were considered (under the assumption that muscle spindles giving muscle length and speed are the ``primary proprioceptors''). Population coding was employed to encode information from robot arm and head joints while it was observing its hand in front of the face and then fed into a standard SOM. A number of limitations of this approach are discussed in \cite{HoffmannBednarova2016}---one of them being that the problem seems ill-posed---the intrinsic dimensionality (topology) of body configuration space is too high to be preserved on a 2-dimensional output sheet. 

In this work, we take a different approach. First, with the bigger goal of developing models testing the hypothesis that self-touch configurations may give rise to representations of the body (and skin surface) in space and to reach to stimuli on own body \cite{Hoffmann2017icdl}, we collected data where the humanoid robot touches itself on the face. Second, population coding of inputs was not used here. Third, the MRF setting was applied.

\section{Methods}

\subsection{Humanoid robot and training dataset}

\begin{figure}
\centering
\begin{subfigure}{.2\textwidth}
  \centering \includegraphics[scale=0.9, width=1\linewidth]{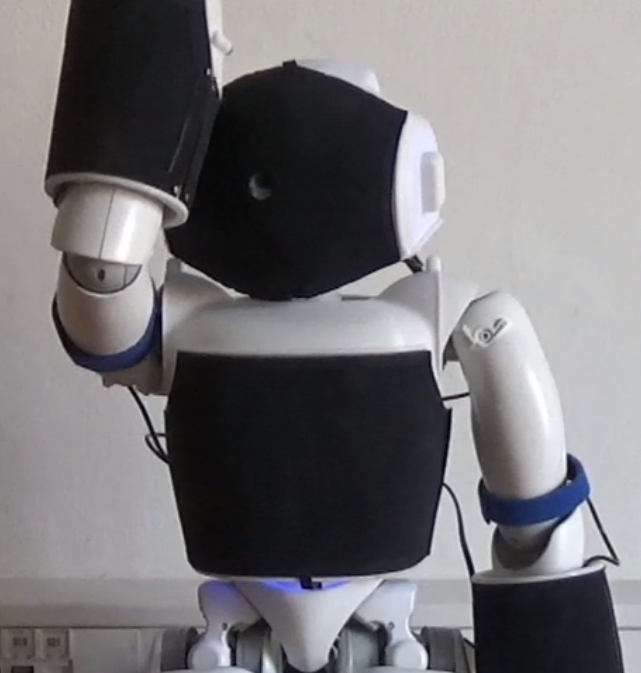}
\end{subfigure}%
\begin{subfigure}{.3\textwidth}
  \centering \includegraphics[width=0.9\linewidth]{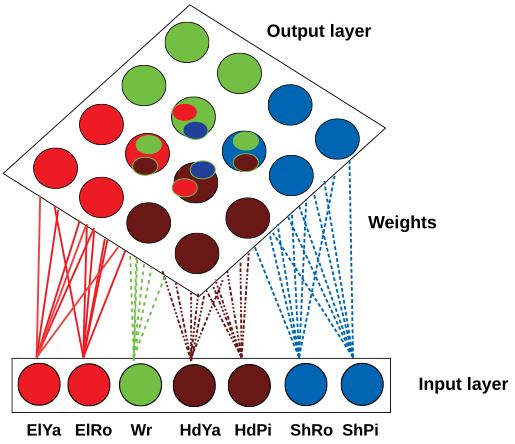}
\end{subfigure}
\caption{(Left) Nao robot with skin performing self-touch. (Right) Overlapping receptive fields setting for MRF-SOM.}
\label{fig:1}
\end{figure}

The training dataset was created by recording the joint angles from 3216 body configurations where the Nao humanoid robot touches its face with its right hand (Fig. \ref{fig:1} left). In every configuration, seven joint angles are recorded: head yaw, head pitch, shoulder roll, shoulder pitch, elbow roll, elbow yaw, and wrist.

\subsection{Self-organizing map and MRF-SOM}

If the dimension of the SOM output layer does not correspond to the intrinsic dimensionality of the input space, distortions of the mapping are inevitable. In \cite{HoffmannStraka2018}, we mapped the artificial skin surface of a humanoid robot (1928 ``taxels'') onto a SOM with 7x24 output neurons. The skin space is locally 2D but enclosed in 3D and there are numerous possible mappings onto a 2D output lattice. By using MRF, we steered the learning into a layout resembling area 3b of the cortex.  

For proprioception, the number of inputs is smaller---here 7 joints (unless population coding is used \cite{HoffmannBednarova2016})---, but the intrinsic dimensionality of the body configuration space seems high: a 2D output sheet cannot preserve all the topological relationships in which postures of the body are ``similar''. Neural data also shows that receptive fields in area 3a typically cover one or two neighbouring joints \cite{Krubitzer2004}. Therefore, we decided to employ the MRF-SOM with a 4x4 hexagonal lattice on the output and 4 partially overlapping receptive fields -- shown in Fig. \ref{fig:1} right.
Distance between neurons is computed using the Manhattan distance, and weights and initial position of neurons are initialised randomly.Ingeniørenes hus møtesenter

\section{Preliminary Results}

The network is evaluated by its ability to encode input joints: individual neuron activation encodes one specific joint or a combination between two or three joints, and topological and pattern encoding where specific activity from a group of neurons encodes a specific input joint or combination.

Results after learning are shown in Fig.~2. The heatmaps show the value of the weight between each neuron of output layer and a corresponding joint. For each cluster (except wrist as there is only one joint), each neuron in the cluster prefers one or another joint, making it possible to differentiate the encoded joints from the state of the cluster. However, despite preferring one joint, most neurons actually encode a combination of the two joints, with high weights values for both. The wrist is also learned by two neurons in its cluster, while another neuron takes it as inhibition (negative weight). Concerning the neurons with overlapping joints from different body parts, it seems that all joints are learned equally: the neuron doesn't show any preference for a specific joint or body part, bur only encodes a combination of all the joints in its receptive field.

\begin{figure}[!ht]
    \label{im:results}
    \centering \includegraphics[scale=0.27]{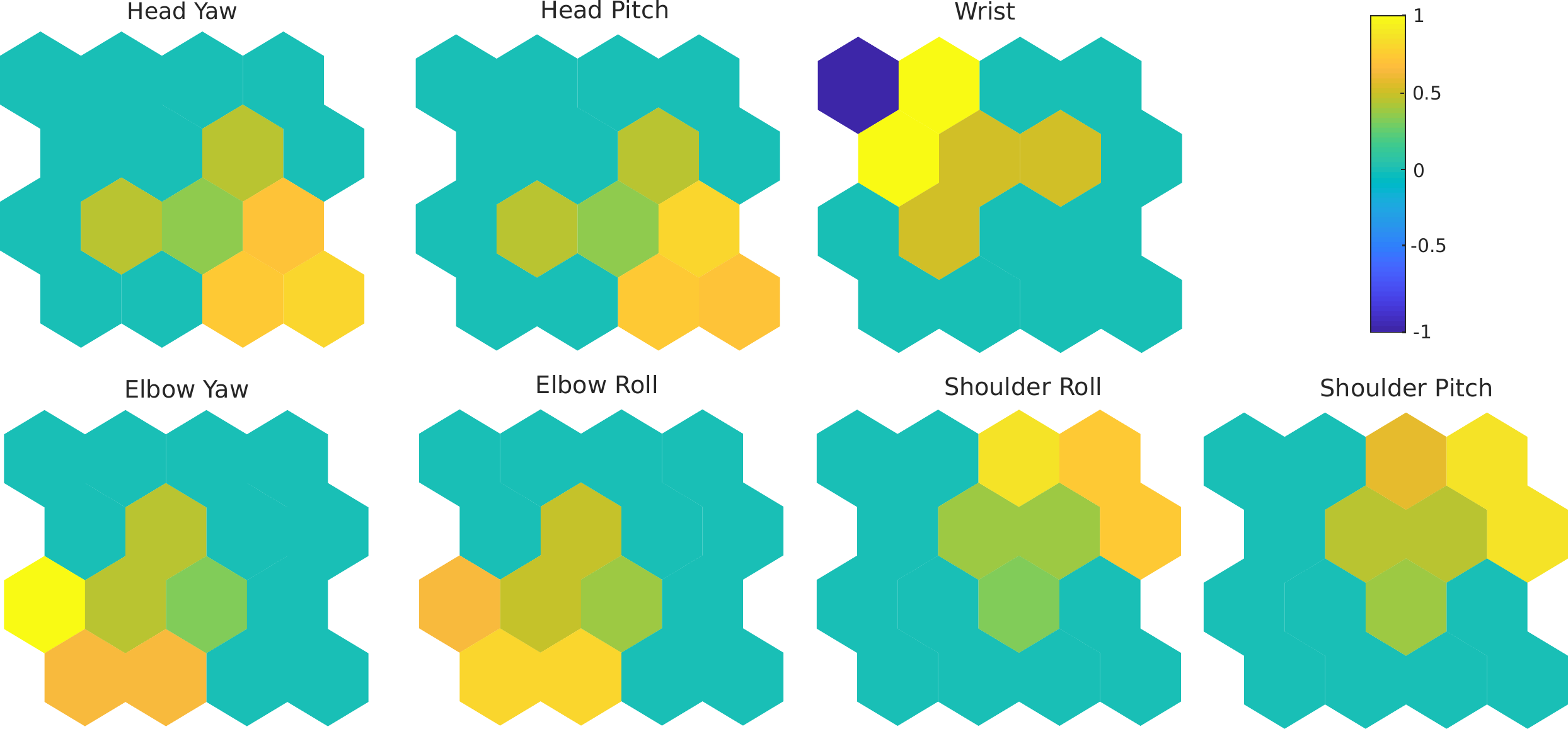}
    \caption{Weight values between input joints and the output layer}
\end{figure}

Another visualization of the network (not shown) displays the distance between neurons. The clusters coding the shoulder and elbow joints have half of the distance between them than with the head or wrist clusters. The wrist cluster is separated from the other joints, while neurons inside the head cluster are extremely close to each other. This indicates that elbow and shoulder joints are more closely related and dependent in the dataset than other joints, while the wrist joint is more independent and head joints are highly correlated together.

\section{Conclusion}

Following up on our previous work \cite{HoffmannBednarova2016,HoffmannStraka2018}, we continue to seek how a robot may learn a representation of its joint space resembling the representations of proprioception found in the somatosensory cortex. A variant of the SOM---MRF-SOM---can be a useful tool to channel the learning in the desired directions. 
However, the results presented are highly preliminary. Even if MRF-SOM can be used to form receptive fields somewhat resembling those in area 3a, it is still not clear what information about the joints is encoded in these areas (see the \textit{position-scaled} vs. \textit{posture-selective} neuron types reported in \cite{Kim2015}) and how it can contribute to the formation of spatial body representations discussed in \cite{Hoffmann2017icdl}.


\bibliographystyle{IEEEtran}
\bibliography{homunculusProprioception_ICDL}

\end{document}